\documentclass[runningheads]{llncs}
\usepackage[T1]{fontenc}
\usepackage{graphicx}
\usepackage{multirow}
\usepackage{booktabs}

\def \ie {i.e., }
\def \eg {e.g., }

\renewcommand{\abstractname}{}

\begin{document}

\title{Exploring text-to-image generation for historical document image retrieval}
%
%


\author{Melissa Cote \and Alexandra Branzan Albu}

%
\institute{University of Victoria, Victoria, British Columbia, Canada\\
\email{\{mcote,aalbu\}@uvic.ca}}

\maketitle              
\begin{abstract}

\keywords{AI text-to-image generation \and document image retrieval \and historical documents \and query-by-example \and visual attributes.}
\end{abstract}

\section{Background including aims}
\label{bg}

Document image retrieval (DIR) aims to identify documents relevant to a query within a set of documents stored as images, based on their content. DIR is essential for managing ever-growing collections of digitized documents, in particular for historical document archives. Current DIR paradigms include textual, layout-based, and query-by-example (QBE) searches. The latter offer the most flexibility in terms of contents to look for; however, they require a sample query document on hand that may not be available. Attribute-based document image retrieval (ABDIR) \cite{cote2024attribute} was recently proposed in that context of non-availability of query documents, to offer users a flexible DIR paradigm based on memorable visual features of document contents. From a computer vision viewpoint, attributes are intermediate level descriptions of objects related to visual qualities \cite{feris2017introduction}, for instance \textquotedblleft spotted\textquotedblright{} when describing a cheetah. For historical documents, examples of attributes can be related to visual content such as \textquotedblleft is illuminated\textquotedblright{} (\ie containing painted flourishes using precious metals), to the condition aspect such as \textquotedblleft has marked deterioration\textquotedblright{} (\ie holes, tears/rips, etc.), or to the document format such as \textquotedblleft has a book format\textquotedblright{}. ABDIR describes document images with attributes determined via convolutional neural network (CNN)-based binary classifiers; these descriptions are combined to form queries of variable complexity, allowing for the retrieval of ranked lists of document images.

In this abstract, we present an exploratory study of the use of generative AI to bridge the gap between QBE and ABDIR. We hypothesize that text-to-image (T2I) generation can be leveraged to create document images using text prompts based on ABDIR-like attributes; the generated document images can then be used as queries within the QBE paradigm. We focus on historical documents as a use case for their diversity and uniqueness in visual features. To the authors' knowledge, this is the first attempt at utilizing T2I generation for DIR.

\section{Methods}
\label{methods}

The typical QBE flowchart starts with a user-selected query document image input to the system (\textquotedblleft step 1: query document image input\textquotedblright{}). The system extracts features from the input to use as a document representation (\textquotedblleft step 2: feature extraction\textquotedblright{}). It then compares the query features to those of the document images in the dataset (extracted offline), using a similarity measure (\textquotedblleft step 3: similarity measurement\textquotedblright{}). The document images in the dataset are then ranked according to their similarity to the query, and the most relevant are retrieved and displayed to the user (\textquotedblleft step 4: ranking and retrieval\textquotedblright{}).

Here, we modify step 1 to create the input query from a text prompt mentioning the desired attributes using T2I generation. We selected Leonardo.Ai \cite{LEONARDOAI2025} as T2I generator for its affordability and high level of customization. The prompts include a rough description of the document type and list the desired attributes. We use the following settings in Leonardo.Ai: Phoenix 1.0 model, no prompt enhancement, dynamic style, medium contrast, \textquotedblleft quality\textquotedblright{} generation mode, small images with a 1:1 or 2:3 ratio, and \textquotedblleft negative\textquotedblright{} option whenever we want to force the absence of an attribute. A prompt example corresponding to the \textquotedblleft has marked deterioration\textquotedblright{} (positive) and \textquotedblleft has wax seal\textquotedblright{} (negative) attributes in ABDIR would be: \textquotedblleft a full page of a historical document that is full of handwritten text and has marked deterioration such as tears/rips, and deep stains\textquotedblright{}, with the negative prompt \textquotedblleft wax seal\textquotedblright{}. For step 2, we follow an approach similar to DeepCBIR \cite{maji2021cbir} and use the activations from the last pooling/averaging layer in a given pre-trained CNN; CNN features are popular as they tend to capture complex semantic information. In our experiments run in Matlab R2023b, we compare the same eight CNN architectures tested in \cite{cote2024attribute}: ResNet-101, DenseNet-201, Inception-v3, Xception, Inception-ResNet-v2, Darknet-53, NASNet-Mobile, and EfficientNet-B0. For step 3, we compare the same similarity measures as in \cite{maji2021cbir}, \ie L1 and L2, and add a third popular measure: cosine distance. As those are dissimilarity measures, in step 4 we rank the document images from the least to the most dissimiliar.

\section{Results}
\label{results}

For straightforward comparisons with ABDIR \cite{cote2024attribute}, we utilized the same set of attributes (\eg \textquotedblleft as wax seal\textquotedblright{}) and queries (combination of positive/present and negative/absent attributes), as well as the same test dataset of 1000 historical document images from HisIR19 \cite{christlein2019icdar}. We report (see Graphics section) the precision@3, @10, and @25, and the R-precision (which adjusts for the number of relevant documents, here ranging from 87 to 346 depending on the query), averaged over all attribute-based queries from \cite{cote2024attribute}, comparing the performance of the various CNN architectures as feature extractors and of the various similarity measures for the proposed T2I-QBE. EfficientNet-B0 is the best tested architecture for T2I-QBE; when combined with L2, it yields the highest precision@3 (0.857±0.178) and precision@10 (0.714±0.135) on average, and when combined with the cosine distance, the highest precision@25 (0.634±0.234). Darknet-53 follows as the second-best architecture, with the highest R-precision on average (0.491±0.135). This compares with 1.000±0.000, 1.000±0.000, 0.966±0.075, and 0.792±0.126 obtained by ABDIR \cite{cote2024attribute} for precision@3, @10, @25, and R-precision, respectively. In addition, the cosine distance yields the best overall performance across architectures and metrics. We show sample qualitative results (see Graphics section) by the proposed T2I-QBE (EfficientNet-B0 and L2) for the first two queries from \cite{cote2024attribute}, where the top retrieved results are all relevant attribute-wise.

\section{Conclusions}
\label{conc}

The experimental results of this exploratory study confirm our hypothesis that T2I generation can be leveraged effectively to create query document images from text prompts based on ABDIR-like attributes and used successfully within the QBE paradigm. EfficientNet-B0, combined with L2, is the best architectural choice. Although its quantitative performance was below that of ABDIR, T2I-QBE appears to be a viable option for historical DIR. One advantage of T2I-QBE over traditional QBE is that the user can refine the prompt as needed to get exactly what they want for the query image, and potentially use several query images with variations on the same attributes by asking the generator for several images at once. Future works will look at extending T2I-QBE to other categories of documents such as academic papers and architectural plans.

\begin{credits}

\end{credits}

%

\section*{Graphics}
\label{graph}

\begin{table*}
\scriptsize
\caption{Performance evaluation of various CNN architectures as feature extractors and of various (dis)similarity measures for the proposed T2I-QBE on the HisIR19 test set of ABDIR \cite{cote2024attribute}. The precision values are averaged over the seven attribute-based queries from \cite{cote2024attribute}. The best value for each metric in shown in bold underlined font.}
\label{tab1}
\setlength{\tabcolsep}{5.7pt}
\renewcommand{\arraystretch}{1.05}
\begin{tabular}{|l|c|c|c|c|c|}
\hline
\multirow{2}{*}{Architecture} & Sim. & \multicolumn{4}{|c|}{Metric (Avg ± Std)} \\
\cline{3-6}
& Measure & Prec@3 $\uparrow$ & Prec@10 $\uparrow$ & Prec@25 $\uparrow$ & R-Prec $\uparrow$ \\
\hline\hline
\multirow{3}{*}{ResNet-101} & L1 & 0.429 ± 0.371 & 0.500 ± 0.200 & 0.446 ± 0.207 & 0.374 ± 0.157 \\
 & L2 & 0.429 ± 0.252 & 0.600 ± 0.289 & 0.503 ± 0.240 & 0.400 ± 0.135 \\
 & Cosine & 0.524 ± 0.262 & 0.643 ± 0.310 & 0.577 ± 0.238 & 0.452 ± 0.128 \\
 \hline
 \multirow{3}{*}{DenseNet-201} & L1 & 0.333 ± 0.193 & 0.443 ± 0.276 & 0.366 ± 0.250 & 0.333 ± 0.187 \\
 & L2 & 0.524 ± 0.178 & 0.457 ± 0.230 & 0.417 ± 0.261 & 0.349 ± 0.172 \\
 & Cosine & 0.524 ± 0.262 & 0.529 ± 0.229 & 0.549 ± 0.234 & 0.433 ± 0.173 \\
 \hline
 \multirow{3}{*}{Inception-v3} & L1 & 0.476 ± 0.262 & 0.429 ± 0.180 & 0.463 ± 0.194 & 0.370 ± 0.075 \\
 & L2 & 0.524 ± 0.262 & 0.443 ± 0.215 & 0.474 ± 0.194 & 0.371 ± 0.075 \\
 & Cosine & 0.619 ± 0.356 & 0.500 ± 0.271 & 0.491 ± 0.255 & 0.387 ± 0.155 \\
 \hline
 \multirow{3}{*}{Xception} & L1 & 0.429 ± 0.418 & 0.457 ± 0.244 & 0.480 ± 0.230 & 0.393 ± 0.124 \\
 & L2 & 0.524 ± 0.378 & 0.514 ± 0.297 & 0.549 ± 0.231 & 0.435 ± 0.096 \\
 & Cosine & 0.619 ± 0.405 & 0.586 ± 0.313 & 0.520 ± 0.289 & 0.438 ± 0.119 \\ \hline
 \multirow{3}{*}{IncResNet-v2} & L1 & 0.429 ± 0.317 & 0.500 ± 0.306 & 0.480 ± 0.276 & 0.370 ± 0.133 \\
 & L2 & 0.429 ± 0.371 & 0.471 ± 0.256 & 0.480 ± 0.280 & 0.371 ± 0.130 \\
 & Cosine & 0.476 ± 0.378 & 0.529 ± 0.320 & 0.497 ± 0.290 & 0.395 ± 0.159 \\\hline
 \multirow{3}{*}{Darknet-53} & L1 & 0.714 ± 0.405 & 0.586 ± 0.324 & 0.589 ± 0.256 & 0.463 ± 0.153 \\
 & L2 & 0.619 ± 0.356 & 0.629 ± 0.309 & 0.606 ± 0.258 & 0.468 ± 0.160 \\
 & Cosine & 0.571 ± 0.371 & 0.643 ± 0.331 & 0.594 ± 0.312 & \underline{\textbf{0.491}} ± 0.186 \\
 \hline
 \multirow{3}{*}{NASNet-Mob} & L1 & 0.381 ± 0.356 & 0.386 ± 0.367 & 0.371 ± 0.295 & 0.322 ± 0.176 \\
 & L2 & 0.429 ± 0.371 & 0.400 ± 0.370 & 0.389 ± 0.276 & 0.305 ± 0.160 \\
 & Cosine & 0.381 ± 0.300 & 0.386 ± 0.339 & 0.417 ± 0.260 & 0.322 ± 0.107 \\\hline
 \multirow{3}{*}{EffNet-B0} & L1 & 0.714 ± 0.230 & 0.600 ± 0.163 & 0.560 ± 0.113 & 0.446 ± 0.077 \\
 & L2 & \underline{\textbf{0.857}} ± 0.178 & \underline{\textbf{0.714}} ± 0.135 & 0.594 ± 0.167 & 0.462 ± 0.074 \\
 & Cosine & 0.714 ± 0.300 & 0.700 ± 0.245 & \underline{\textbf{0.634}} ± 0.234 & 0.447 ± 0.136 \\
\hline
\end{tabular}
\end{table*}

\begin{figure}
\includegraphics[width=\textwidth]{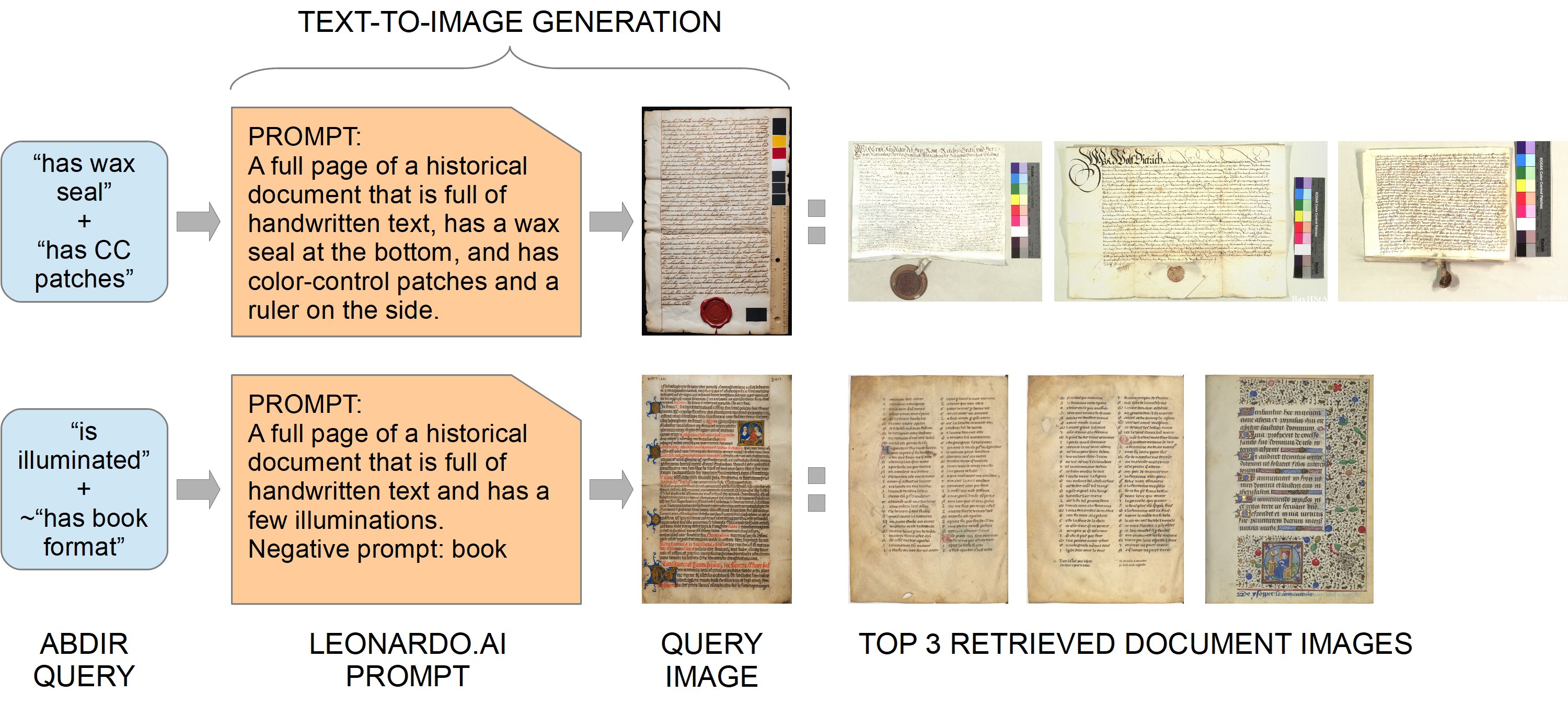}
\caption{Sample historical document image retrieval results for the first two queries from \cite{cote2024attribute} by the proposed T2I-QBE (EfficientNet-B0 for feature extraction and L2 for (dis)similarity measurement). From left to right: original attribute-based query from ABDIR, prompt for T2I generation, query document image, top 3 retrieved results. }
\label{fig1}
\end{figure}

\end{document}